\pdfoutput=1

\documentclass[11pt]{article}

\usepackage[]{acl}

\usepackage{times}
\usepackage{latexsym}
\usepackage{amsmath, amssymb, amsfonts}
\usepackage{graphicx}
\usepackage{makecell}
\usepackage[OT2, T1]{fontenc}
\usepackage[russian, english]{babel}
\usepackage[T1]{fontenc}

\usepackage[utf8]{inputenc}

\usepackage{microtype}

\usepackage{inconsolata}

\title{MALTO at SemEval-2024 Task 6: Leveraging Synthetic Data for LLM Hallucination Detection}

\author{
    Federico Borra$^1$\thanks{These authors contributed equally to this work.}  \\\And
    Claudio Savelli$^1$\footnotemark[1]  \\\And
    Giacomo Rosso$^1$ \\\AND
    Alkis Koudounas$^1$ \\ \And
    Flavio Giobergia$^1$ \\ \\
    \textsuperscript{1} Politecnico di Torino, Italy \\ \And
  }
  
\begin{document}
\maketitle
\begin{abstract}
In Natural Language Generation (NLG), contemporary Large Language Models (LLMs) face several challenges, such as generating fluent yet inaccurate outputs and reliance on fluency-centric metrics. 
This often leads to neural networks exhibiting ``hallucinations.'' 
The SHROOM challenge focuses on automatically identifying these hallucinations in the generated text. 
To tackle these issues, we introduce two key components, a data augmentation pipeline incorporating LLM-assisted pseudo-labelling and sentence rephrasing, and a voting ensemble from three models pre-trained on Natural Language Inference (NLI) tasks and fine-tuned on diverse datasets.

\end{abstract}



\section{Introduction}
Natural Language Generation (NLG) models are AI systems that use neural networks to produce human-like text.
They have shown significant advancements in recent years, particularly with the advent of transformer-based architectures such as GPT (Generative Pre-trained Transformer) \cite{radford2018improving}.
These models offered unprecedented levels of fluency and coherence in generated text \cite{HAN2021225}.
However, a critical challenge arises: these models can produce linguistically fluent but semantically inaccurate outputs, a phenomenon referred to as \textit{hallucination}~\cite{Ji_2023}. 

To address this challenge, the Shared-task on Hallucinations and Related Observable Overgeneration Mistakes (SHROOM) has been proposed at SemEval 2024.
In particular, the Shared task aims to address the existing gap in assessing the semantic correctness and meaningfulness of NLG models.
The ever-increasing adoption of such models makes it necessary to automatically detect and mitigate semantic hallucinations \cite{huang2023survey}.

Some examples to tackle hallucination detection tasks in literature \cite{Ji_2023} are:
(i) \textit{Information Extraction and Comparison} between a generated text and a ground truth,
(ii) \textit{Natural Language Inference Metrics} that express the entailment between generated text and a ground truth or
(iii) \textit{Faithfulness Classification Metrics} that leverage upon knowledge-grounded datasets.

In this work, we address the SHROOM shared task by introducing an automatic pipeline of hallucination detection through the comparison between a generated text and a ground truth text.
We propose enriching the original data available using different augmentation techniques, including LLM-aided pseudo-labeling and sentence rephrasing.
Additionally, we suggest using an ensemble of three different approaches, incorporating a simple BERT-based classifier, a model trained through Conditioned Reinforcement Learning Fine Tuning (C-RLFT)~\cite{wang2023openchat}, and a sequential model based on iterative fine-tuning.
We show how this ensemble benefits from using different, complementary approaches, in particular in terms of recall. 
Our methodology obtained an accuracy of 80.07\% in the SemEval-Task 6 SHROOM. 
 
\subsection{Dataset}
\label{sec:dataset}
The dataset available for the SHROOM challenge is a collection of objects.
Each object represents a solution of a generative language model to either of three tasks: \textit{Definition Modeling} (DM), \textit{Machine Translation} (MT), and \textit{Paraphrase Generation} (PG).
Each solution has been annotated, based on its contents, as either an \textit{hallucination} of the generative model or \textit{not hallucination} by 5 human annotators. 

For each object, the available information includes (i) the \textit{source (src)}, which is the input text given to the generative language model, (ii) the \textit{hypothesis (hyp)}, which represents the generated textual output of the model, and (iii) the \textit{target (tgt)} which is the intended reference or ``gold'' text that the model is supposed to generate. 
Additionally, the task field indicates the type of task being solved, either DM, MT, or PG. The label, either `\textit{`hallucination''} or \textit{``not hallucination''}, is determined through majority voting among five annotators, with \textit{p(hal)} indicating the proportion of annotators who labeled the data point as a hallucination.


The gold (and augmented) data cardinalities are defined in Table~\ref{tab:augmentation}. 
The training dataset comprises 500 instances with gold labels, denoted as $\mathcal{D}_{g}$, and 30,000 unlabelled instances, referred to as $\mathcal{D}_{u}$ (10,000 for each of the three tasks). 
The evaluation split contains 1,500 labelled samples, with 500 instances used for validation ($\mathcal{D}_{v}$) and 1,000 for testing ($\mathcal{D}_{t}$).
We use the validation set for fine-tuning the ensemble layer (refer to Section~\ref{sec:ensemble}), while the final test set provides overall results (see Section~\ref{sec:results}).

We further rephrase the original 500 labelled sentence of the training set ($\mathcal{D}_{r}$ in the table, see Section~\ref{subsec:sentencerephrasing}), while applying weak labelling the 30,000 unlabelled instances ($\mathcal{D}_{pl}$, see Section~\ref{subsec:pseudolabeling}).


\section{Methodology}
\begin{figure*}[!ht]
    \centering
    \includegraphics[width=\linewidth]{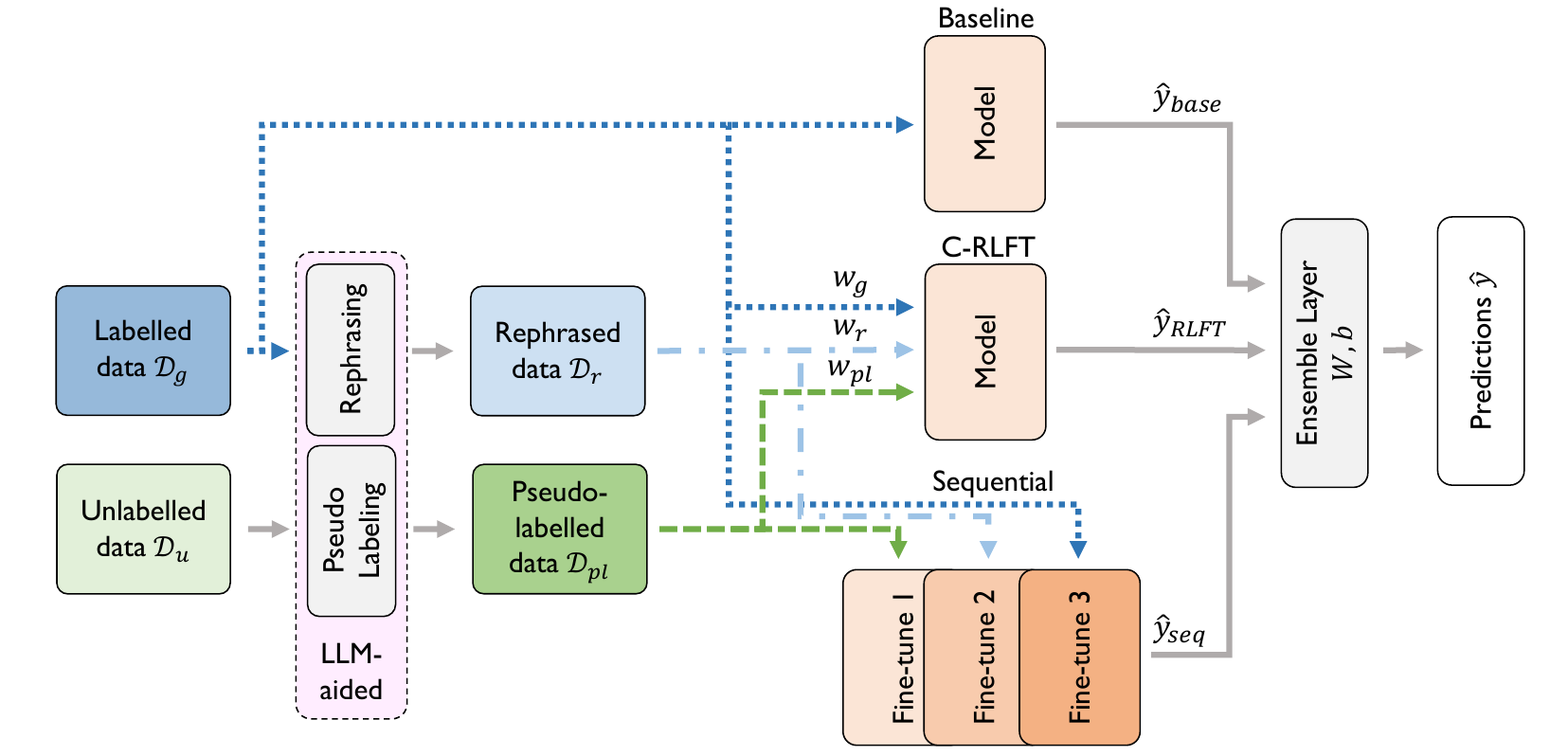}
    \caption{Pipeline architecture depicting data augmentation techniques and weighted ensemble of strategies}
    \label{fig:architecture}
\end{figure*}

The main goal of this work is to propose a binary classification model to predict whether the answer to a given query is a hallucination or not.
Figure \ref{fig:architecture} presents the main architecture adopted to address this task\footnote{The code to replicate the experiments and the results will be released upon acceptance.}.
We propose (i) using a data augmentation pipeline (see Section~\ref{sec:augmentation}) consisting of Large Language Model (LLM)-aided pseudo-labelling and sentence rephrasing and
(ii) adopting an ensemble model (see Section~\ref{sec:ensemble}) based on the results of three models, defined as follows:

\begin{itemize}
    \item \textit{Baseline} model, a binary classifier based on a semantic-aware embedding (e.g. BERT-based \cite{devlin2019bert}).
    The baseline model is presented in Section \ref{sec:baseline}
    \item \textit{C-RLFT} (Conditioned Reinforcement Learning Fine Tuning \cite{wang2023openchat}), based on the introduction of pseudo-labels and augmented data, with different weighting schemes based on the quality of each data point.
    We cover C-RLFT in more detail in Section \ref{sec:crlft}
    \item \textit{Sequential} model, based on the iterative fine-tuning of the baseline model with increasingly higher-quality data, as detailed in Section \ref{sec:seq}
\end{itemize}


\subsection{Data Augmentation}
\label{sec:augmentation}
Due to the scarcity of data, we developed an approach to extend the number of labelled samples we could use to train our models.
We specifically leverage two distinct techniques: pseudo-labelling and sentence rephrasing. Both approaches are based on LLMs and, as such, may themselves be subject to hallucinations or inaccuracies. As detailed next, we mitigate this problem by (1) using the C-RLFT technique~\cite{wang2023openchat}, which involves assigning different weights to mixed-quality samples, and (2) with a sequential training that introduces different-quality labels at different training stages.

\subsubsection{Pseudo Labeling}\label{subsec:pseudolabeling}
As stated in Section \ref{sec:dataset}, only a small fraction of the dataset available is labelled. We introduce additional pseudo labels, as obtained by querying an LLM in a few-shot learning setting. Based on the hardware available, we tested several LLM models to assess the reliability of the pseudo labels produced (in terms of accuracy). We identified SOLAR~\cite{kim2023solar} as being the best-performing model among the pool of candidates. Thus, we leverage it to generate synthetic labels for unlabelled data through a few-shot learning approach. We refer to this augmented dataset as $\mathcal{D}_{pl}$.


\subsubsection{Sentence Rephrasing}\label{subsec:sentencerephrasing}
We utilized sentence rephrasing based on GPT-4 as an additional data augmentation technique. 
We do so by rephrasing both the model output and the target output of each gold sample.
This approach aims to provide the model with diverse data while maintaining the reliability of the labels. We refer to this dataset as $\mathcal{D}_r$.



 
\subsection{Models}
We adopt an ensemble of three models, as described below. All models are based on DeBERTa~\cite{he2021deberta}. More specifically, we use a baseline model that has been fine-tuned in different ways.

\subsubsection{Baseline}
\label{sec:baseline}
We employed a baseline model utilizing the DeBERTa encoder pre-trained on the Natural Language Inference (NLI) task, with a binary classification head. We fine-tune this model on the provided classification task using only data with the gold labels available, referred to as $\mathcal{D}_g$. 
The training approach involved minimizing the Binary Cross Entropy (BCE) loss.

We use the probability $p(hal)$ as the ground truth instead of the binary label. This is done to better reflect the distribution of votes of the human annotators in the output logits of the model. 




\subsubsection{C-RLFT}
\label{sec:crlft}

Conditioned Reinforcement Learning Fine Tuning (C-RLFT) is a technique that refines models using coarse-grained reward labels, allowing fine-tuning with both expert and sub-optimal data lacking preference labels. 
In our specific scenario, we fine-tuned the model by assigning different weights to data based on their label type, i.e., synthetic or gold. The weight assigned to each data sample influences the contribution to the final BCE loss. 


We define a weighting scheme for the gold dataset $\mathcal{D}_g$, the pseudo-labelled dataset $\mathcal{D}_{pl}$ and the rephrased dataset $\mathcal{D}_r$, as follows:


\begin{equation*}
    w(x_i) =
\left\{ 
    \begin{array}{l}
        w_g \quad \textit{if}\; x_i \in \mathcal{D}_{g}\\
        w_r \quad \textit{if}\; x_i \in \mathcal{D}_{r}\\
        w_{pl} \quad \textit{if}\; x_i \in \mathcal{D}_{pl}\\
    \end{array} 
\right.
\end{equation*}
We choose weights $w_g > w_r > w_{pl}$. In this way, we aim to assign a higher importance to gold labels due to their reliability. The lowest weight is assigned to the pseudo-labelled points because of the lower quality of the automatically assigned labels. An intermediate weight is given to rephrased sentences due to the higher quality of the ground truth w.r.t. the pseudo-labelled points. The weighted loss is thus defined as follows, for a point $x_i$ with ground truth $y_i$, as computed for a binary classifier $f(\cdot)$:

\begin{align*}
wBCE(x_i, y_i) =& -w(x_i) \cdot (y_i\;log\; f(x_i) \\
               & + (1 - y_i)\; \log(1 - f(x_i)))
\end{align*}  

\subsubsection{Sequential}
\label{sec:seq}
The third model used is based on a sequential strategy that uses both generated and augmented data. We introduce three fine-tuning steps, performed sequentially on the initial model. 
The model initially underwent fine-tuning using the pseudo-labelled dataset $\mathcal{D}_{pl}$, which is the lowest-quality dataset among the three available. Subsequently, we fine-tuned the resulting model on the rephrased data $\mathcal{D}_r$, which benefits from the original, correct, labels.
The final fine-tuning step is then executed on the golden truth dataset $\mathcal{D}_{g}$.
This approach is inspired by curriculum learning ~\cite{soviany2022curriculum}, with data being ordered by veracity instead of difficulty.

This strategy aims to enhance the model's understanding of the task by starting with a substantial amount of data, including the less reliable synthetic labels, and progressively updating the model parameters with increasingly consistent data. This sequential approach allows the model to first adapt to the task using a broader dataset and then refine its knowledge with the highest quality data available.

\subsection{Ensemble}
\label{sec:ensemble}
The final step in the proposed pipeline involves creating an ensemble of results from the previously-introduced techniques, which has already proven to be effective in other NLP tasks~\cite{jia2023review, koudounas2023barhotti}. We trained three distinct models (\textit{baseline}, \textit{C-RLFT}, \textit{sequential}) with specific strategies, and we generated their outputs ($\hat{y}_{base}$, $\hat{y}_{RLFT}$, $\hat{y}_{seq}$) on a validation set of previously unseen gold data. We obtain a single result $\hat{y}$ from the previous ones by using a single-layer network ($W \in \mathbb{R}^{3}$ and $b \in \mathbb{R}$), as follows:
\begin{equation}
\label{eq:ensemble}
\begin{aligned}
    &\text{$\hat{y}_{models}$} = (\hat{y}_{base}, \hat{y}_{RLFT}, \hat{y}_{seq}) \\
    &\text{$\hat{y}$} = \sigma (W^\intercal\hat{y}_{models} + b)
\end{aligned}
\end{equation}
This network is trained to predict a single output from the three models' predicted probabilities. We trained this network by minimizing a BCE function.
\section{Experimental Results}
\label{sec:results}
This section presents the experimental setup used, and the main results obtained. 

\subsection{Experimental Setup}
The dataset used to train and validate the model is the one made available in the SHROOM challenge's model agnostic track (refer to Section~\ref{sec:dataset}). The augmentations are specified in Section~\ref{sec:augmentation}. 


\begin{table}
\centering
\begin{tabular}{cccc}
\hline
\textbf{Dataset Type} & \textbf{Label} & \textbf{Split} & \textbf{\#Samples} \\
\hline
$\mathcal{D}_{g}$ & yes & Train & 500 \\
$\mathcal{D}_{r}$ & yes & Train & 500 \\
$\mathcal{D}_{u}$ & no & Train & 30,000 \\
$\mathcal{D}_{pl}$ & weak & Train & 30,000 \\
\hline
$\mathcal{D}_{v}$ & yes & Val & 500 \\
\hline
$\mathcal{D}_{t}$ & yes & Test & 1000 \\
\hline
\end{tabular}
\caption{Dataset type, labelling, and number of instances for each considered split.}
\label{tab:augmentation}
\end{table}



For the model backbone and synthetic labelling we leverage Huggingface pre-trained models\footnote{We use \textit{deberta-xlarge} and \textit{deberta-xlarge-mnli} as encoders, \textit{TheBloke/SOLAR-10.7B-Instruct-v1.0-GPTQ} for pseudo labelling.}. We also leverage \textit{GPT-4} for sentence rephrasing. All the experiments' results are obtained based on 5 different runs. 
For C-RLFT, we identified the best performance for weights $w_g = 1.01, w_r = 0.4, w_{pl} = 0.1$. 

\subsection{Model performance}

We summarize the results obtained on the test set in Table~\ref{tab:res}. We report the results in terms of $F_1$ score, precision and recall on the ``Hallucination'' class, as well as overall accuracy.

\begin{table*}
\centering
\begin{tabular}{cccccc}
\hline
\textbf{Model} & \textbf{Method} & $\mathbf{F_1}$ \textbf{score} & \textbf{Precision} & \textbf{Recall} & \textbf{Accuracy} \\
\hline
DeBERTa & Baseline   & 0.6207$\pm$0.0808             & 0.7112$\pm$0.0661             & 0.5588$\pm$0.1562             & 0.7254$\pm$0.0231 \\
DeBERTa & C-RLFT     & 0.6182$\pm$0.0857             & \underline{0.8081$\pm$0.0939} & 0.5089$\pm$0.1574             & 0.7476$\pm$0.0245 \\
DeBERTa & Sequential & \underline{0.7075$\pm$0.0394} & \textbf{0.8169$\pm$0.0396}    & \underline{0.6253$\pm$0.0690} & \textbf{0.7898$\pm$0.0194} \\
DeBERTa & Ensemble   & \textbf{0.7119$\pm$0.0272}    & 0.7918$\pm$0.0402             & \textbf{0.6474$\pm$0.0466}    & \underline{0.7867$\pm$0.0171} \\
\hline
D.+MNLI & Baseline & 0.7138$\pm$0.0253 & 0.7420$\pm$0.0319 & \textbf{0.6882$\pm$0.0372} & 0.7753$\pm$0.0178 \\
D.+MNLI & C-RLFT & 0.6146$\pm$0.0917 & \textbf{0.8410$\pm$0.0706} & 0.4900$\pm$0.1376 & 0.7528$\pm$0.0302 \\
D.+MNLI & Sequential & \underline{0.7320$\pm$0.0229} & \underline{0.8177$\pm$0.0233} & 0.6628$\pm$0.0329 & \textbf{0.8024$\pm$0.0141} \\
D.+MNLI & Ensemble & \textbf{0.7371$\pm$0.0223} & 0.8016$\pm$0.0347 & \underline{0.6829$\pm$0.0425} & \underline{0.8017$\pm$0.0143} \\
\hline
\end{tabular}
\caption{Performance metrics for DeBERTa and DeBERTa + MNLI models. Best scores are highlighted in bold, second-best are underlined.}
\label{tab:res}
\end{table*}

We use as backbone both DeBERTa and a version of DeBERTa that has been fine-tuned on the Machine Natural Language Inference (MNLI) task. Further discussions on the choice of the backbone are presented in Section~\ref{sec:backbone}.

Section~\ref{sec:strategies} highlights the result differences for each of the considered strategies and includes additional considerations on the ensemble of the approaches. Finally, we provide qualitative examples of the results in Section~\ref{sec:example}.

\subsection{Backbone impact}
\label{sec:backbone}
We start by examining the differences between two backbone models, both fine-tuned on the gold data only -- these are referred to as the \textit{baseline} models in Table~\ref{tab:res}. 

There is a notable increase of 0.09 in the $F_1$ score for the MNLI-fine-tuned model compared to the original DeBERTa. Interestingly, all the proposed DeBERTa-based approaches are still outperformed by the baseline DeBERTa+MNLI-based model (although to a lesser extent). This highlights the close relationship between the tasks of \textit{Hallucination Detection} and \textit{Natural Language Inference}. 


\begin{table*}
\centering
\resizebox{\linewidth}{!}{
\begin{tabular}{c c c c c}
\hline
\textbf{src} & \textbf{hyp} & \textbf{tgt} & \textbf{Target} $\mathbf{p(hal)}$ & $\mathbf{\hat{p}(hal)}$ \\
\hline
\makecell{\foreignlanguage{russian}{Король Харальд} \\ \foreignlanguage{russian}{Гормссон, более} \\ \foreignlanguage{russian}{известный как} \\ \foreignlanguage{russian}{Харальд Синезубый,} \\ \foreignlanguage{russian}{ввёл в Дании} \\ \foreignlanguage{russian}{христианство.}} &
\makecell{King Harald \\Hormsson, better \\ known as \\Harald Sinezubii, \\ introduced Christianity \\ to Denmark.} &
\makecell{King Harald \\Gormsson, better \\ known as \\“Harald Bluetooth", \\ introduced Christianity \\ to Denmark.} & 0.40 & 0.40 \\
\hline
\makecell{Why'd you got \\ to go and do that?} & 
\makecell{Why did you have \\ to go do that?} &
\makecell{Why would you \\ say that?} & 0.00 & 0.91 \\
\hline
\end{tabular}}
\caption{Examples of correctly and wrongly predicted as \textit{``Hallucination''} or \textit{``Not Hallucination''}. The model output is \textit{p(hal)} and must be confronted with gold \textit{p(hal)}. The first example proposed is a Russian to English \textit{Machine Translation} (MT), and the second is an English \textit{Paraphrase Generation} (PG).}
\label{tab:examples}
\end{table*}

\subsection{Strategies comparisons}
\label{sec:strategies}

The \textit{baseline} strategy, which utilizes all available labelled gold data, establishes a lower bound in the expected performance. Both \textit{C-RLFT} and \textit{sequential training} exhibit substantial performance improvements. 

Regarding the \textit{ensemble} strategy, the results in terms of $F_1$ score outperform individual techniques.
We observe a trade-off where the precision of the final result is slightly compromised in exchange for an improved recall. 
This suggests that the \textit{ensemble} effectively identifies instances of hallucination overlooked by the standalone approaches. These advantages are consistent across both backbones implementations, with and without the additional MNLI fine-tuning.

In a setting where detected hallucinations are shown to the final user with a warning, we argue that the recall is a metric of greater interest (w.r.t. precision). A false negative could be potentially harmful since final users are not warned of the presence of possible hallucinations. A false positive would raise a warning that may be inspected by the final user and safely ignored. 

The weights learned for the ensemble layer, based on Equation \ref{eq:ensemble}, are $W = (0.52, 1.7, 1.82)$, $b = -1.7$. This shows how both C-RLFT and the sequential models are weighted similarly and more heavily w.r.t. the baseline. The baseline is assigned a non-zero weight: it is considered, although to a lesser extent, in the final vote. The negative bias implies a learned prior: without further knowledge, the initial prediction is of a negative one (i.e., the majority class). 

\subsection{Qualitative Example}
\label{sec:example}
Table~\ref{tab:examples} demonstrates the effectiveness of the applied strategies through some qualitative examples. 
We specifically showcase the sentences with the minimum (first row) and maximum (second row) errors. 

The first instance depicts a partial hallucination, attributed to the transliteration of ``Sinezubii'' instead of the translation ``Bluetooth,'' which is absent from the translation hypothesis. In the second example, despite a paraphrased similarity between the source and hypothesis, the target introduces an action (``saying'') not present in the source (``doing''). As such, we argue that this might be a case of incorrectly labelled ground truth. 



\section{Conclusions}
This work tackles the SHROOM Task 6 challenge at SemEval 2024, focusing on semantic hallucination in NLG models. 
We propose an automatic pipeline for hallucination detection, utilizing data augmentation and an ensemble of three different methodologies. 
The ensemble of the approaches obtained an accuracy of 80.07\% in the task's leaderboard.
Particular attention should also be paid to the results obtained with the novelty method \textit{sequential}, which was able to outperform the results of the other two methods due to the proposed sequential training.



\bibliography{custom}




\end{document}